\documentclass[12pt]{article}
\usepackage{mathrsfs}
\usepackage{stmaryrd}
\usepackage{bbding}
\usepackage{amssymb}
\usepackage{subfigure}
\usepackage{pifont}
\usepackage{amsmath}
\usepackage{longtable}
\usepackage{tabularx}
\usepackage{stmaryrd}
\usepackage{setspace}
\usepackage[dvips]{color}
\usepackage{framed}

\newcommand{\ba}{\begin{array}}
\newcommand{\ea}{\end{array}}
\newcommand{\be}{\begin{equation}}
\newcommand{\ee}{\end{equation}}
\newcommand\bes{\begin{eqnarray}}
\newcommand\ees{\end{eqnarray}}

\newcommand{\vc}[1]{{\mbox{\boldmath${#1}$}}}

\newcommand{\Rmnum}[1]{\expandafter\@slowromancap\romannumeral #1@}
\usepackage[dvips]{graphics,color}
\usepackage{graphicx}
\usepackage{caption}

 \makeatother
\begin{document}
\date{}
\baselineskip 15pt
\title{{\bf{Panel semiparametric quantile regression neural network for electricity consumption forecasting}}}
\author{{Xingcai Zhou, Jiangyan Wang}\\ \\
{\emph{Institute of Statistics and Data Science, Nanjing Audit University}}\\ {\emph{Nanjing, 211815, China}}\\\\
}
\maketitle
\begin{center}
\begin{minipage}{6in}
\textbf{Abstract:} China has made great achievements in electric power industry during the long-term deepening of reform and opening up. However, the complex regional economic, social and natural conditions, electricity resources are not evenly distributed, which accounts for the electricity deficiency in some regions of China. It is desirable to develop a robust electricity forecasting model. Motivated by which, we propose a Panel Semiparametric Quantile Regression Neural Network (PSQRNN) by utilizing the artificial neural network and semiparametric quantile regression. The PSQRNN can explore a potential linear and nonlinear relationships among the variables, interpret the unobserved provincial heterogeneity, and maintain the interpretability of parametric models simultaneously. And the PSQRNN is trained by combining the penalized quantile regression with LASSO, ridge regression and backpropagation algorithm. To evaluate the prediction accuracy, an empirical analysis is conducted to analyze the provincial electricity consumption from 1999 to 2018 in China based on three scenarios. From which, one finds that the PSQRNN model performs better for electricity consumption forecasting by considering the economic and climatic factors. Finally, the provincial electricity consumptions of the next $5$ years (2019-2023) in China are reported by forecasting.
\end{minipage}
\end{center}

\vskip 0.3cm
\onehalfspacing
\section{Introduction}
Over the past $40$ years of reform and opening up, China's energy industry has undergone tremendous changes and made remarkable achievements. The total amount of energy production and consumption ranks first in the world. Moreover, the clean energy consumption structure has been greatly optimized, and the total amount of which stands in the front ranks, see \cite{CEPPE2018}. Energy development has injected a continuous impetus into social and economic development, among which, the electricity, an important secondary energy source, has been widely used and become the material basis of modern civilization since it is easy to clean and control. By the end of $2017$, the installed capacity of electric power in China had reached to $1.778$ billion kilowatts (kW), among the highest number in the world.

However, the electricity resources are not evenly distributed given that China has a vast territory and complex societal, economic and natural conditions, which is reflected in Figure 1, the Annual electricity consumption distributions of $1999$, $2005$, $2010$ and $2017$ in China.  In which, the local or regional electricity shortages are highlighted, for example, in 2002, 21 provinces suffered from power shortages; the severest electricity shortage (for about 30 million kW) occurred in 2011 since 2004, which is twice as much as the total power generation of Anhui province, see \cite{Yuetal2015}. Electricity is also concerned to be a vital driver of economic development. A robust and accurate forecasting model is desirable for policy maker in both developed and developing countries. Consequently, helping to capture the future development trend of economy and the electricity demand, the annual Electricity consumption forecasting (ECF) is imperative.
ECF plays a critical role in monitoring and planning the transportation of electric power. In addition, it helps to save energy, reduce pollution, and improve the security and stability of the power system. But ECF is highly affected by many factors such as population, economy growth, industrial structure, income level of residents, climate, national policy, power facilities and so on. It makes ECF a challenging task.

\begin{center}
\begin{figure}
 \centerline{ \includegraphics[height=10cm, width=14cm]{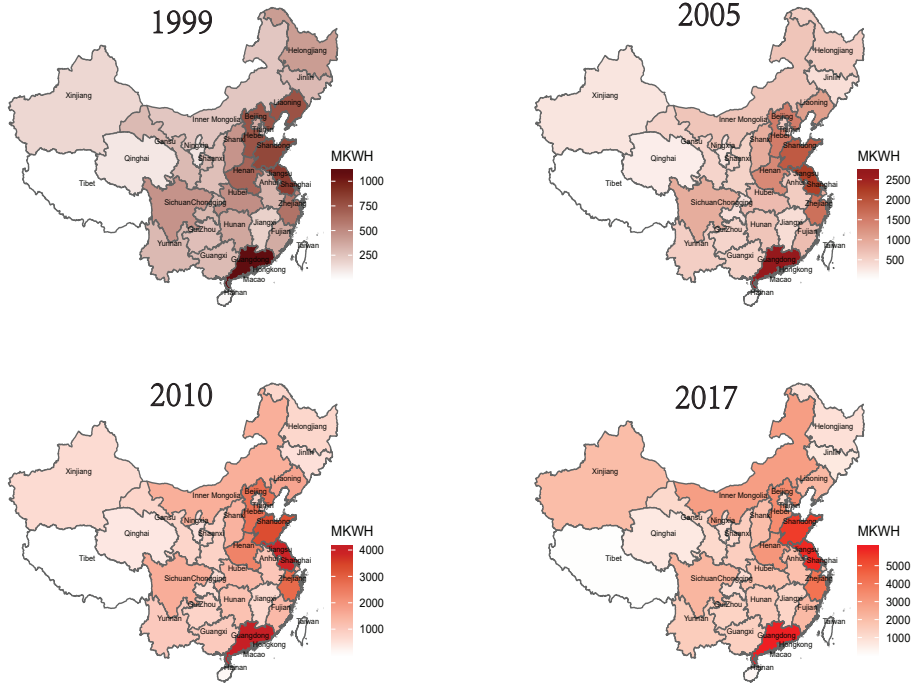}}
  \vskip -0.5cm
  \caption{Annual electricity consumption distributions of 1999, 2005, 2010 and 2017 in China.}\label{figure1}
\end{figure}
\end{center}

\vskip -1.5cm

Recent years have witnessed the development of various forecasting techniques, which can be divided into four major groups: autoregressive integrated moving average (ARIMA), multivariate linear regression (MLR), grey prediction models and artificial intelligence (AI) learning. As one of the most popular time series models, ARIMA is widely used for ECF.
Ediger and Akar \cite{Ediger2007} applied ARIMA to predict primary energy demand of fuel in Turkey.
Mohamed et al. \cite{Mohamedetal2010} considered double seasonal ARIMA model to forecast load demand in Malaysia.
Filik et al. \cite{Filik2011} proposed a hourly forecast of long-term electric energy demand by ARMA in Turkey.
Xu et al. \cite{Xuetal2015} introduced a grey model and ARMA (GM-ARMA) method to predict the energy consumption in China.
Hussain et al. \cite{Hussainetal2016} applied Holt-Winter and ARIMA to forecast electricity consumption in Pakistan.
Cabral et al. \cite{Cabraletal2017} presented a spatial ARIMA model for ECF in Brazil.
Oliveira and Oliveira \cite{Oliveira2018} forecasted mid-long term electricity consumption by combining ARIMA with exponential smoothing and bootstrap aggregating technique.
MLR is a statistical approach which is widely used for ECF, see for instance, Bianco et al. \cite{Biancoetal2013}, Meng and Niu \cite{MengNiu2011} and Kaytez et al. \cite{Kaytezetal2015}. This predictor can be carried out easily, but the flexibility and fitting of MLR and ARIMA may not be satisfactory in practice because of the pre-defined constraint.
Grey prediction model is also a commonly used tool in electricity consumption prediction, details can be found in Wang et al. \cite{Wangetal2018}, Ding et al. \cite{Dingetal2018}, Tang et al. \cite{Tangetal2019}. As far as we know, AI methods are the highly popular prediction tools, which is able to perform well in dealing with modeling complexity and nonlinearity.
Different AI methods have been proposed for electricity consumption/demand/load prediction, for example, Artificial Neural Networks (ANNs) (Azadeh et al.\cite{Azadehetal2008a}, Azadeh et al. \cite{Azadehetal2008b}, Kandananond \cite{Kandananond2011}, He et al. \cite{Heetal2019}), Support Vector Machine (SVM) (Pai and Hong \cite{PaiHong2005}, Fan and Chen \cite{FanChen2006}, Hong \cite{Hong2009}), Support vector regression (SVR) (Elattar et al. \cite{Elattaretal2010}, Yang et al. \cite{Yangetal2019}) and Genetic programming (GP) (Mostafavi et al. \cite{Mostafavietal2013}), among others.
However, these techniques work well only in time series or cross-sectional data, not the panel data.

In this paper, we propose a novel approach called Panel Semiparametric Quantile Regression Neural Network (PSQRNN) to analyze and forecast the provincial electricity consumption in China by ANN and semiparametric quantile regression (SQR). Quantile regression (QR) of Koenker and Bassett \cite{KoenkerBassett1978} explores the relationship among variables comprehensively, including median regression as a special case. Since quantile estimator can quantify the entire conditional distribution of the response variable conditional on covariates and give a global assessment of the covariate effects at different quantile of the response (Koenker \cite{Koenker2005}).
It is particularly useful when the conditional distribution is heterogeneous and asymmetric, heavy-tailed, or truncated in \cite{Shimetal2012}.
However, most research of QR (Koenker\cite{Koenker2005}) with multiple predictors have relied on linear or parametric nonlinear models, see Cannon \cite{Cannon2011}.
To take the advantage of QR and ANN, Taylor \cite{Taylor2000} introduced a more flexible model: Quantile Regression Neural Network (QRNN),
which can implement a nonlinear QR and grasp nonlinear relationship between the response and covariates without specifying a precise functional form. Related work includes Xu et al. \cite{Xuetal2016}, He and Li \cite{HeLi2018}, He et al. \cite{Heetal2019}. Further, Xu et al. \cite{Xuetal2017} used a composite quantile regression neural network (CQRNN) model to explore the potential nonlinear relationship among variables. Cannon \cite{Cannon2018} considered non-crossing nonlinear regression quantiles by monotone CQRNN, called MCQRNN; Jiang et al. \cite{Jiangetal2017} developed an expectile regression neural network by adding ANN to expectile regression, which is similar to QRNN. Since QRNN model is a purely nonlinear QR, the flexibility and good forecasting performance is guaranteed. However, QRNN is considered as a black-box model due to the weakness of interpretability.

Recently, semiparametric regression model has become popular because it keeps the flexibility of nonparametric models and maintains the interpretability of parametric models simultaneously (Kai et al. \cite{KaiLiZou2011}). The application of semiparametric model is dated back to Engle et al. \cite{Engleetal1986}, Fan and Hyndman \cite{FanHyndman2012}, Weron and Misiorek \cite{WeronMisiorek2008}, Shao et al. \cite{Shaoetal2014}, Goude et al. \cite{Goudeetal2014}, Shao et al. \cite{Shaoetal2015}. In the background of semiparametric quantile regression (SQR), Lebotsa et al. \cite{Lebotsaetal2018} considered a short term electricity demand forecasting using partially linear additive QR.

We focus on the heterogeneity of the provincial electricity consumption based on a cross-province study. In conditional mean panel data (linear) models, taking a difference is commonly used to eliminate the individual effect, but it is invalid in the background of QR, particularly for linear QR model. But such study is rare in literatures (Cai et al. \cite{Caietal2018}).
Koenker \cite{Koenker2004} first introduced a panel quantile regression (PQR) which treats the individual fixed effects as a pure location shift parameters common to all conditional quantiles.
Later, Lamarche \cite{Lamarche2010} contributed on studied theoretical properties of PQR, whereas Calvao \cite{Galvao2011} extended the quantile regression to a dynamic panel data model with fixed effects.
The application of PQR includes: Chen and Lei \cite{ChenLei2018} revisited the environment-energy-growth nexus by employing a PQR to incorporate the effects of renewable energy consumption and technological innovation; Wang et al. \cite{WangZhuetal2018} investigated the effect of democracy, political globalization, and urbanization on PM 2.5 concentrations with G20 countries based on evidence from PQR; Wang et al. \cite{WangZengLiu2019} applied a PQR and a balanced city panel data in China to examine the multiple impacts of technological progress on CO$_2$ emission. However, such PQR models are linear panel quantile regression. To address the issue of nonlinearities and heterogeneity simultaneously, Cai et al. \cite{Caietal2018} proposed a semiparametric quantile panel data model with correlated random effects, in which some of the coefficients are allowed to depend on smooth economic variables while the other coefficients are constant, to estimate the growth effect of foreign direct investment,
where the nonparametric component is used to model nonlinearity, but it is not flexible enough compared to neural network.
Moreover, nonparametric nonlinear model suffers from model misspecification, which leads to an inaccurate forecast.
Motivated from which, we propose the PSQRNN to forecast the provincial panel electricity consumption in China.

The contributions of our paper are: (1) PSQRNN is proposed for the regional electricity consumption forecasting, which combines panel data, semiparametric model and composite QR with QRNN.
(2) PSQRNN is a new framework by adding parametric structure and unobserved regional heterogeneity to QRNN, which can explore potential linear
and nonlinear relationship among the variables and interpret the unobserved cross-sectional heterogeneity simultaneously, and maintain a better interpretability of parametric models.
(3) A new estimator is derived to train the PSQRNN model by assembling the penalized quantile regression with LASSO, ridge regression and backpropagation algorithm.

The rest of the paper is organized as follows. In Section 2, we introduce the PSQRNN model and address the issues of corresponding estimation and selection. Section 3 gives the comparison of PSQRNN with three competitive intelligent methods including BP neural network (BP), SVM and QRNN, and presents the provincial electricity consumption forecasting in China under three scenarios via training, evaluating and forecasting. Conclusions are drawn in Section 4.

\section{Panel Semiparametric Quantile Regression Neural Network}

\subsection {Model setup}   
\subsubsection {Linear quantile regression and QRNN}
To explore the relationship between a response $y$ and predictors $X=(x_1,\cdots,x_p)^T$ for a time series and cross sectional data, the linear quantile regression (LQR) model can be written as
\be
Q_\tau(y_i|\vc X_i)=\vc X_i^T\vc\beta_\tau,
\ee
where $Q_\tau(y_i|\vc X_i)$ is the $\tau$th condition quantile of $y_i$, $\{(\vc X_i,y_i), i=1,\cdots,n\}$ is the observed data, $\vc X_i=(x_{i1},\cdots, x_{ip})$, $n$ is sample size, $\beta_\tau$ is regression coefficients, and $\tau\in(0,1)$, which follows the settings of Koenker \cite{Koenker2005}. This model only considers the linear relationship between the response variable
and the predictors under different quantiles, which is inappropriate to investigate the nonlinear relationship in practice. Thus, Cannon \cite{Cannon2011} proposed a QRNN model in term of one hidden layer,
\be
Q_\tau(y_i|\vc X_i)=f\left(\sum_{j=1}^Jg_j(i)w_j^{(o)}+b^{(o)}\right),
\ee
with
\be
g_j(i)=a\left(\sum_{k=1}^px_{ik}^Tw_{kj}^{(h)}+h_j^{(h)}\right),
\ee
where $w_j^{(h)}$ and $b^{(h)}$ are the hidden-layer weights and bias, $w_j^{(o)}$ and $b^{(o)}$ are the output-layer weights and bias, and $J$ is the number of node of the hidden layer, $a(\cdot)$ is the active function and $f(\cdot)$ is the output-layer transfer function. One can refer to the schematic diagram in Figure 2 (a), which shows a QRNN model with two hidden layers.

\subsubsection{Panel semiparametic quantile regression}
In this paper, we study a panel data of provincial electricity consumption. Let a scalar dependent variable $Y_{it}$ be the observation of $i$th individual at time $t$ for $i=1,\cdots N$ and $t=1,\cdots,T$. The panel semiparametric quantile regression model (PSQR) is:
\be\label{PSQR}
Q_\tau(Y_{it}|\vc X_{it},\vc Z_{it},U_{it},\alpha_i)=\vc Z_{it}^T\vc \beta_\tau+\vc X_{it}^T\vc\gamma_\tau(U_{it})+\alpha_i,
\ee
where $Q_\tau(Y_{it}|\vc X_{it},\vc Z_{it},U_{it},\alpha_i)$ is the $\tau$th quantile of $Y_{it}$ given $\vc Z_{it}, \vc X_{it}, U_{it}$ and $\alpha_i$, $\vc Z_{it}$ and $\vc X_{it}$ are $p\times1$ times $q\times 1$ predictors, respectively. $\vc\beta_\tau$ is a constant regression coefficient, $\vc\gamma_\tau(U_{it})$ denotes a regression functional coefficient of $U_{it}$, which is also an observable scale predictor, and $\alpha_i$ is an individual effect. See Cai et al. \cite{Caietal2018}.

\begin{center}
\begin{figure}
 \centerline{ \includegraphics[height=10cm, width=14cm]{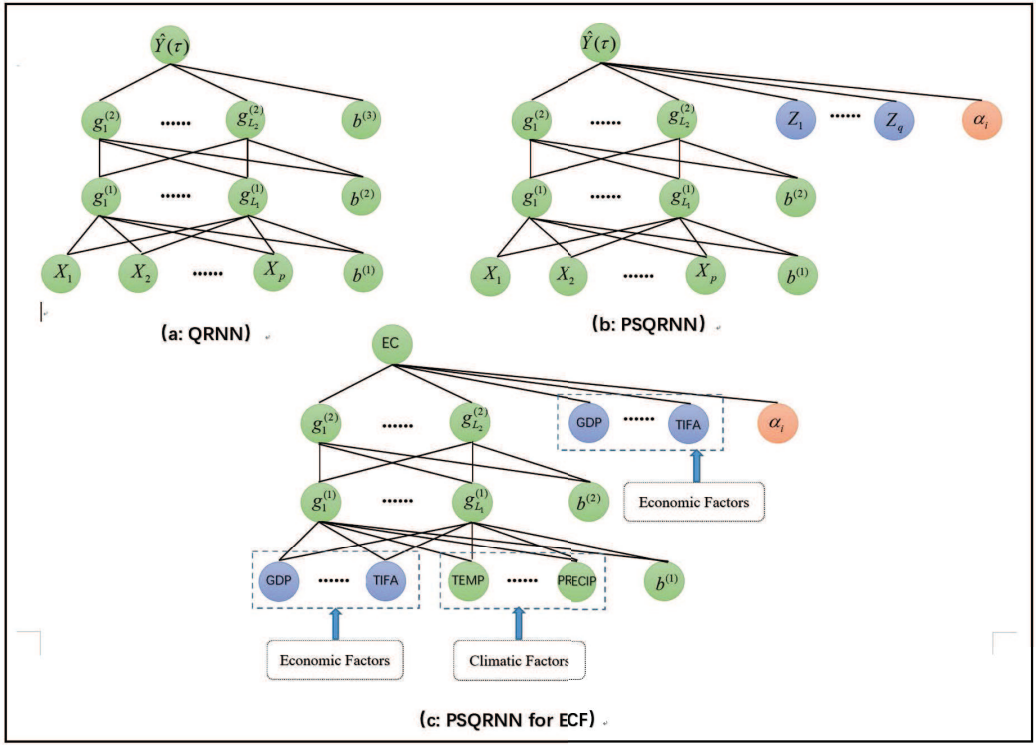}}
  \caption{The frameworks of  (a) QRNN, (b) PSQRNN and (c) PSQRNN for ECF}\label{figure2}
\end{figure}
\end{center}

\vskip -1.5cm

The model PSQR (\ref{PSQR}) is a traditional statistical model, which combined LQR for panel data (Koenker \cite{Koenker2004}) and nonparametric regression model. To avoid the ``curse of dimension" in pure nonparametric model, a varying coefficient model is embedded into the PSQR. The varying coefficient $\vc X_{it}^T\vc \gamma_\tau(U_{it})$ in model (\ref{PSQR}) concerns the nonlinear relationship of the response $Y$ and the predictors $(\vc X, U)$. The nonparametric nonlinear form and predictions are forcibly divided into $\vc X$ and $U$, with a high risk of model misspecification. In order to avoid an inaccurate forecasting, we propose a new panel semiparametric quantile regression model based on an artificial neural network.

\subsubsection {PSQRNN}
Motivated by LQR and QRNN, we develop a new learning method to predict provincial electricity consumption in China, which combines PSQR and ANN. We call it PSQRNN, which is designed under a multi-layer perceptron (MLP) framework. To begin with, a general dependent structure of PSQRNN is:
\be\label{psqrnn}
Q_\tau(Y_{it}|\vc X_{it},\vc Z_{it},\alpha_i)=\vc Z_{it}^T\vc\beta_\tau+ANN(X_{it};{\vc\theta}_\tau)+\alpha_i,
\ee
where $ANN(\cdot)$ is an arbitrary nonlinear function, whose structure is simulated by utilizing ANN, and $\vc\theta_\tau$ is a parameter of weights and biases in ANN. Let the input and output layer is the $0$th layer and the $L+1$th layer, $\vc g_{it}^{(0)}=\vc X_{it}$, $\vc g_{it}^{(l)}=(g_{1,it}^{(l)},g_{2,it}^{(l)},\cdots,g_{n_l,it}^{(l)})^T$ is a vector of $n_l$ nodes at the $l$th hidden layer for $l=1,\cdots,L$, i.e., $\vc g_{it}^{(l)}\in R^{n_l\times1}$, while the weighted matrix $\vc W^{(l)}\in R^{n_{l-1}\times n_l}$ and the hidden-layer bias $\vc b^{(l)}\in R^{n_l\times1}$.

First, at the input layer,
\be
\vc g_{it}^{(1)}=a^{(1)}\left({\vc W^{(1)}}^T\vc X_{it}+\vc b^{(1)}\right).
\ee

At the hidden layer,
\be
\vc g_{it}^{(l)}=a^{(l)}\left({\vc W^{(l)}}^T\vc g_{it}^{(l-1)}+\vc b^{(l)}\right), l=2,\cdots,L,
\ee
where $a^{(l)}(\cdot)$ is an activation function, which controls the network's nonlinearity, maps the real line to its subset, and often adopts forms like signmoid($\cdot$), tanh($\cdot$), ReLU($\cdot$), softplus($\cdot$), etc.

At the output layer, an estimator of the $\tau$th conditional quantile of response is then given by
\be\label{hatpsqr}
\hat{Q}_\tau(Y_{it}|\vc X_{it},\vc Z_{it},\alpha_i)=f\left(\vc Z_{it}^T\vc\beta_\tau+{\vc W^{(L+1)}}^T\vc g_{it}^{(L)}+\alpha_i\right),
\ee
where $f(\cdot)$ is the output-layer transfer, which is usually an identity function. Here, $ANN(\cdot)={\vc W^{(L+1)}}^T\vc g_{it}^{(L)}$.
The schematic diagram is depicted in Figure 2 (b), which shows a PSQRNN model with two hidden layers.

\subsection {Model estimation}  
In this section, we first introduce a penalized quantile regression model for panel data, then the estimation procedure of PSQRNN is proposed.
\subsubsection {Estimation of the penalized quantile regression}
In model (\ref{psqrnn}), if the nonlinear $ANN(\cdot)==0$, the PSQRNN becomes a linear quantile panel data model,
\be
Q_\tau(Y_{it}|\vc X_{it},\vc Z_{it},\alpha_i)=\vc Z_{it}^T\vc\beta_\tau+\alpha_i,
\ee
which is proposed by Koenker \cite{Koenker2004}, where $\alpha_i$ is treated as fixed effects with pure location shift effects on the conditional quantiles of the response variable, but the effects of regressions depends on the quantile. While in penalized quantile regression, the quantile loss function is minimized by adding $\ell_1$ penalty on fixed effects. The objective function is:
\be\label{eq_pqr}
\max_{\vc\alpha,\vc\beta}\sum_{k=1}^K\sum_{i=1}^N\sum_{t=1}^Tw_k\rho_{\tau_k}\left(Y_{it}-\alpha_i-\vc X_{it}^T\vc\beta_{\tau_k}\right)+\lambda\sum_{i=1}^N|\alpha_i|,
\ee
where $\rho_\tau(u)=u(\tau-I(u<0))$ is the quantile loss function. The weights $w_k$ controls the $K$ quantiles $\{\tau_1,\cdots,\tau_q\}$ when estimating parameters $\alpha_i$, and $\lambda$ is the tuning parameter which reduces the individual effects to improve the accuracy and robustness of the estimation of $\beta$. In terms of which, Lamarche \cite{Lamarche2010}, Galvao \cite{Galvao2011} and Canay \cite{Canay2011} studied the penalized quantile regression model for panel data (\ref{eq_pqr}) from different perspectives.

\subsubsection{Estimation of PSQRNN}
Inspired by the estimation of PQRPD in (\ref{eq_pqr}) and the implementation of QRNN in Cannon \cite{Cannon2011}, we define the loss function of PSQRNN as
\begin{eqnarray}
\mathcal{L}(\theta ) &=&\frac{1}{KNT}\sum_{k=1}^{K}\sum_{i=1}^{N}%
\sum_{t=1}^{T}w_{k}\rho _{\tau _{k}}\left( Y_{it}-\hat{Q}_{\tau _{k}}(Y_{it}|%
{\mbox{\boldmath${X}$}}_{it},{\mbox{\boldmath${Z}$}}_{it},\alpha
_{i})\right)   \notag \\
&&+\lambda _{1}\frac{1}{N}\sum_{i=1}^{N}|\alpha _{i}|+\lambda _{2}\frac{1}{%
N_{L}}\sum_{l=1}^{L}\sum_{i=1}^{n_{l-1}}\sum_{j=1}^{n_{l}}\left(
w_{ij}^{(l)}\right) ^{2} \notag
\end{eqnarray}
where $\hat{Q}_{\tau_k}(Y_{it}|\vc X_{it},\vc Z_{it},\alpha_i)$ is given in (\ref{hatpsqr}), $\theta=\left\{\alpha_i,W^{(l)},b^{(l)}|i=1,\cdots,N,l=1,\cdots,L\right\}$, $W^{(l)}=(w_{ij}^{(l)})_{n_{l-1}\times n_l}$, $N_L=\sum_{l=1}^Nn_{n-l}n_l$, and $\lambda_1$, $\lambda_2$ are the model penalty parameters, respectively. Here, we apply $\ell_1$ shrinkage to individual effects $\alpha_i$ and conventional Gaussian $\ell_2$ penalties to weights $w_{ij}^{(l)}$. To improve the accuracy of prediction, $\lambda_1$ shrinks the individual effects estimators toward zero and $\lambda_2$ effectively prevents the model from overfitting.

Typically, the loss $\mathcal{L}(\theta)$ with weights and biases of ANN is minimized by gradient descent (GD) and backpropagation algorithm. However, the check functions $\rho_\tau(\cdot)$ and $|\cdot|$ usually are non-differentiable, which is obvious since the derivative is not valid at the origin. Instead, one can replace $\rho_\tau(\cdot)$ and $|\cdot|$ with approximations that are differentiable everywhere. Following Chen \cite{Chen2007} for quantile regression and Cannon \cite{Cannon2011} for QRNN, the Huber norm, which provides a smooth transition between absolute and squared errors around the origin, is defined as follows:
\be
h^\varepsilon(u)=\left\{\begin{array}{ccl}
               \frac{u^2}{2\varepsilon}, &if&  0\leq |u|\leq\varepsilon,\\
               |u|-\frac{\varepsilon}{2}, &if& |u|>\varepsilon,
             \end{array}\right.
\ee
which is employed to approximate $\rho_\tau(\cdot)$ and $|\cdot|$ by
\be
\rho_\tau^\varepsilon(u)=\left\{\begin{array}{ccl}
               \tau h^\varepsilon(u), &if& u\geq0,\\
               (\tau-1)h^\varepsilon(u), &if& u<0,
             \end{array}\right.
\ee
and $|\alpha_i|=h\varepsilon(\alpha_i)$,
where $\varepsilon$ is a pre-deterministic threshold, with $\varepsilon=2^{-i}$ for $i=-8,-9,\cdots,-32$, which is default in the R package \texttt{qrnn}. Hence, the standard GD optimization algorithm is employed to optimize the approximate loss function in the following
\begin{eqnarray}
\mathcal{L}^{\varepsilon }(\theta ) &=&\frac{1}{KNT}\sum_{k=1}^{K}%
\sum_{i=1}^{N}\sum_{t=1}^{T}w_{k}\rho _{\tau _{k}}^{\varepsilon }\left(
Y_{it}-\hat{Q}_{\tau _{k}}(Y_{it}|{\mbox{\boldmath${X}$}}_{it},{%
\mbox{\boldmath${Z}$}}_{it},\alpha _{i})\right)   \notag \\
&&+\lambda _{1}\frac{1}{N}\sum_{i=1}^{N}h(\alpha _{i})+\lambda _{2}\frac{1}{%
N_{L}}\sum_{l=1}^{L}\sum_{i=1}^{n_{l-1}}\sum_{j=1}^{n_{l}}\left(
w_{ij}^{(l)}\right) ^{2}.  \label{Alossfuc}
\end{eqnarray}
This procedure is conducted by using \texttt{nlm} for Newton-type algorithm or \texttt{optim} for Nelder-Mead, and quasi-Newton, conjugate-gradient algorithms in R package. In the entire optimization procedure, $\varepsilon$ in $\mathcal{L}^\varepsilon(\theta)$ begins with a larger starting value, and is updated in each iteration. The algorithm converges until $\varepsilon$ goes to zero.

\subsection{Model selection} 
The PSQRNN model is flexible and useful to reveal the nonlinear predictor-predicted relationship, and contains some latent interactions between predictors. The complexity of ANN in PSQRNN model is determined by $(p,L,n_1,\cdots,n_L)$. A model that is too complex may result in over-fitting, but can be avoided by penalizing a larger weight in the input-hidden layer by adding a quadratic penalty term, which has been considered in our model. $\lambda_2$ in (\ref{Alossfuc}) contributes to the terms of weight decay.

Another issue of PSQRNN modeling is to choose $(p,L,n_1,\cdots,n_L)$ and $(\lambda_1,\lambda_2)$, which play an important role in training and prediction. In practice, $p$ is a fixed parameter since predictors in ANN are predetermined. However, to find an optimal combination is unrealistic due to the computational consuming. In practice, $L$ is taken to be $1$ or $2$. As a criterion for model selection, the Bayesian information criterion (BIC) is applied in this paper. For $L=1$, one defines  YQ$_{it1}$=$Y_{it}-\hat{Q}_{\tau _{k}}(Y_{it}|{\mbox{\boldmath${X}$}}%
_{it},{\mbox{\boldmath${Z}$}}_{it},\alpha _{i};n_{1},\lambda _{1},\lambda
_{2})$, and
\begin{eqnarray}
&&BIC_{1}(n_{1},\lambda _{1},\lambda _{2})=\ln \left( \frac{1}{KNT}%
\sum_{k=1}^{K}\sum_{i=1}^{N}\sum_{t=1}^{T}w_{k}\rho _{\tau
_{k}}^{\varepsilon }\text{YQ}_{it1}\right)   \notag \\
&&+\frac{1}{2}\frac{\ln (NT)}{NT}[(p+2)n_{1}+q+N].
\end{eqnarray}
The optimal values of hyperparametrics $(n_1, \lambda_1,\lambda_2)$, $(n_1^*, \lambda_1^*,\lambda_2^*)$ are determined by
\be
(n_1^*, \lambda_1^*,\lambda_2^*)=\mathbf{argmax}_{(n_1, \lambda_1,\lambda_2)}BIC_1(n_1, \lambda_1,\lambda_2).
\ee

For $L=2$, one defines
\begin{eqnarray}
&&BIC_{2}(n_{1},n_{2},\lambda _{1},\lambda _{2})=\ln \left( \frac{1}{KNT}%
\sum_{k=1}^{K}\sum_{i=1}^{N}\sum_{t=1}^{T}w_{k}\rho _{\tau
_{k}}^{\varepsilon }\text{YQ}_{it2}\right)   \notag \\
&&+\frac{1}{2}\frac{\ln (NT)}{NT}[(p+1)n_{1}+n_{2}(n_{1}+2)+q+N],
\end{eqnarray}
where YQ$_{it2}$=$Y_{it}-\hat{Q}_{\tau _{k}}(Y_{it}|{\mbox{\boldmath${X}$}}%
_{it},{\mbox{\boldmath${Z}$}}_{it},\alpha _{i};n_{1},n_{2},\lambda
_{1},\lambda _{2})$, and
\be
(n_1^*, n_2^*,\lambda_1^*,\lambda_2^*)=\mathbf{argmax}_{(n_1, n_2, \lambda_1, \lambda_2)}BIC_2(n_1, n_2, \lambda_1, \lambda_2).
\ee
In general, grid search method can be used to minimize $BIC$, which is simply an exhaustive searching method through a manually specified subset of the hyperparameter space. In the empirical analysis, we adopt two hidden layers, that is, $L=2$. The four-dimensional gird search method for the case of $L=2$ is time-consuming. In reality, $n_1$ and $n_2$ are predesigned, and a two-dimensional grid search method is used to choose an optimal $(\lambda_1^*,\lambda_2^*)$. We find that the PSQRNN model with a small number of nodes fits and predicts well, when $L=2$.

\subsection{Key notes of the model training and sketch}
A sketch of training and prediction is provided in the following.

$\bullet$ The selection of $K$, $\tau_k$ and $w_k$. In (\ref{eq_pqr}) and (\ref{Alossfuc}), fitting multiple values of $\tau$ simultaneously allows one to ``borrow strength" from regression quantile and improve the global model performance (Cannon \cite{Cannon2018}). Typically, $\tau_k=k/(k+1)$, $k=1,\cdots,K$ are equally spaced. $w_k$ are weights that allow regression quantiles for each $\tau_k$ to contribute to the total error. Constant weights $w_k=1/K$ yield a standard composite quantile regression error function in (\ref{eq_pqr}) and (\ref{Alossfuc}). We set $w_k=1/K$ since there is no prior information. Koenker \cite{Koenker2004} pointed out that the choice of $w_k$ and $\tau_k$ is analogous to the choice of discretely weighted $L$-statistics, see for instance, Mosteller \cite{Mosteller1946}. $K$ is generally selected by experience such as $3$, $5$ and $9$. For a seriously skewed data, $K$ is selected to be $15$ or $20$.

\begin{center}
\begin{figure}
 \centerline{ \includegraphics[height=16cm, width=13cm]{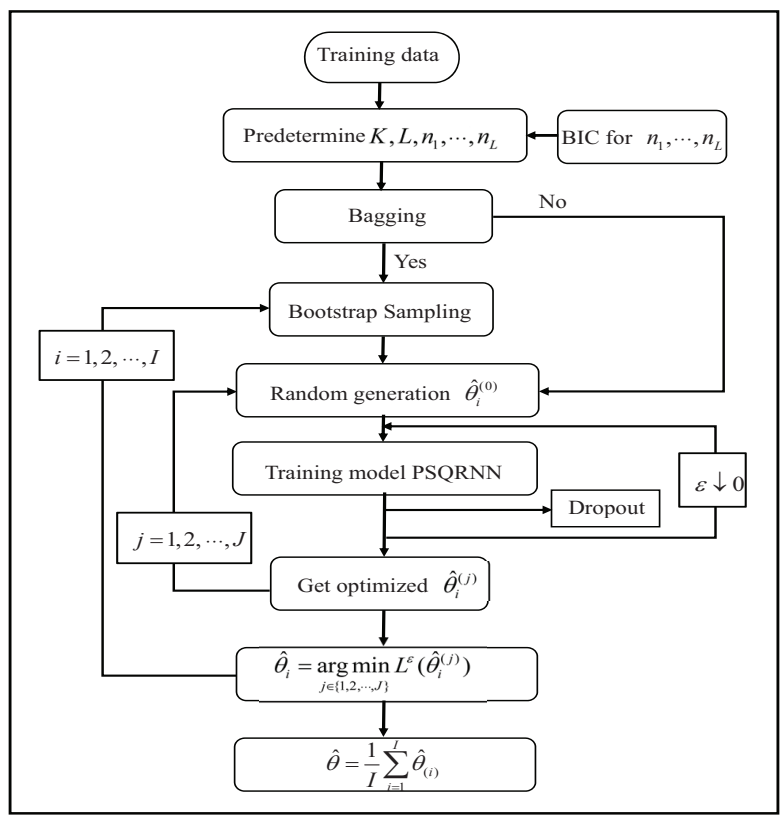}}
 \vspace{-1in}
  \caption{Sketch of the PSQRNN model training.}\label{figure3}
\end{figure}
\end{center}

\vskip -1.1cm

$\bullet$ The selection of $a^{(l)}(\cdot)$ and $f(\cdot)$. The activation function of the hidden layer $a^{(l)}(\cdot)$ yields the nonlinearity of the network, which is usually taken to be $\mathrm{sigmoid}(x)=1/(1+e^{-x})$, $\mathrm{tanh}(x)=(e^x-e^{-x})/(e^x+e^{-x})$, $\mathrm{softplus}(x)=\log(1+e^x)$ and $\mathrm{ReLU}(x)=\max(0,x)$. ReLU is the most popular activation function for neural networks, and is successfully applied to deep neural networks. The greatest advantage of ReLU is to alleviate the vanishing gradient problem (Clevert et al. \cite{Clevertetal2016}). Recently, Leaky ReLU (LReLU), which is a variant of ReLU, has been shown to be superior to ReLU (Maas et al. \cite{Maasetal2013}). In our model training, the activation function is $a^{(l)}(x)=\mathrm{ELU}(x)$ for classification with $\alpha=1$ for $l=1,\cdots,L$, and the output-layer transfer $f(x)=x$ for regression.

$\bullet$ Avoid local minima and saddle points. $\mathcal{L}^\varepsilon(\theta)$ in (\ref{Alossfuc}) is a continuous non-convex loss function. Minimizing a non-convex objective function is a big challenge for science and engineering. Gradient descent or quasi-Newton methods are ubiquitously used to carry out such minimizations. There are many techniques to avoid local minima and saddle points (Dauphin et al. \cite{Dauphinetal2014}), specially in the deep neural networks. In this paper, we repeat to train the PSQRNN model by generating different weights and bias as initial parameters, then take the minimum point.


The sketch of the model training is depicted in Figure \ref{figure3}.

\setcounter{equation}{0}
\section{Provincial electricity consumption forecasting in China}
\subsection{Data}
Many works have explored the influencing factors of electricity consumption, which can be roughly divided into economic factors (He et al. \cite{Heetal2019}) and climatic factors (Fan et al. \cite{Fanetal2019}). The impact of the two factors regarding to electricity consumption is extensively studied. However, few of them pay attention to the nonlinear case. In this paper, PSQRNN is applied to analyze and forecast China's electricity consumption. We choose economic factors as Gross Regional Product (GDP), Value-added of Secondary Industry (VASI), Total Retail Sales of Consumer Goods (TRSCG), Total Imports and Exports (TIE), and the climatic factors as Annual Average Temperature (AAT), Annual Average Relative Humidity (AARH), Days of Precipitation ($\geq 0.1mm$) per year (DP) and Sunshine Hours (SH). The panel data of China's $30$ provinces during 1999-2018 is collected, in which dataset of EC, GDP, VASI, TRSCG and TIE are from National Bureau of Statistics of China: {\color{blue} http://www.stats.gov.cn/tjsj/ndsj}; dataset of AAT, AARH, DP and SH are from National Meteorological Information Center of China: {\color{blue} http://data.cma.cn}. Among which, the data of economic factors are complete, but the data of climate factors are missing, so we deal with it by imputation via interpolating the mean. The dataset used in PSQRNN model is summarized in Table \ref{table1}.

{\setlength{\tabcolsep}{1.6em}{
\begin{table}[!h]
\begin{center}
\caption{\bf  Variables used in PSQRNN model}\label{table1}
\vskip 0.1cm
\tabcolsep 0.2cm
\renewcommand{\arraystretch}{1.2}
{\begin{tabular}{cccccc}
  \hline
  Variable  & Description & Unit & Obs. & Missing (\%) & Type \\
  \hline
  EC         & Electricity consumption & $10^8$ KWH   &    30$\times$20&    0.00&  Both \\
  GDP      & Gross regional product & $10^8$ Yuan  &   30$\times$20&    0.00&    Both\\
  VASI      & Value-added of secondary industry&  $10^8$ Yuan  &    30$\times$20&    0.00&    Both \\
  TRSCG  &Total retail sales of consumer goods& $10^8$ Yuan    &   30$\times$20&    0.00&   Both \\
  TIE         & Total imports and exports&  $10^8$ USD &  30$\times$20  &    0.00 &   Both\\
  AAT       & Annual average temperature   & $^oC$ &  30$\times$20  &  0.33    & nonparametric\\
  AARH    & Annual average relative humidity   & \% &  30$\times$20  &  0.33   & nonparametric\\
  DP         & Days of precipitation in a year  & Days &  30$\times$20  & 3.00  & nonparametric\\
  SH         &  Sunshine hours in a year  & Hours & 30$\times$20  & 4.67   & nonparametric\\
  \hline
\end{tabular}
}
\end{center}
\emph{Note: ```Obs." denotes the observations; `30$\times$20" stands for $30$ provinces times $20$ years (1999-2018); `Both" means that a variable appears both in the parametric linear and nonparametric neural network of the PSQRNN model.}
\end{table}}}

We do some descriptive statistical analysis of the dataset, and find that
(i) These panel data are complex with many inherent or latent tends; (ii) Based on Skewness, Kurtosis and p-value of Jarque-Bera test per year, the distributions of EC, GDP, VASI, TRSCG and TIE are skewed (Skewness$>0$), which is more concentrated than normal distribution with longer tails (Kurtosis$>3$). This indicates the non-normality of the unconditional distribution of these factors (p-value$<0.05$). Thus a quantile regression method is desirable to describe the heterogeneity of the factors in electricity consumption. (iii) The distributions of AAT, AARH,
DP and SH per year can not reject the normality, but there are great differences among provinces. In addition, the five indicators increase and the variation tends to be larger year by year; TIE deviates much more from the nominal level, which also poses a difficulty for prediction.

\subsection {Comparison of prediction methods}
In this subsection, the performance of the PSQRNN for ECF is investigated by empirical analysis. Of interest is the comparison of PSQRNN with three competitive intelligent methods including BP neural network (BP), SVM and QRNN. To evaluate the prediction, the provincial EC and the provincial influence factors (GDP, VASI, TRSCG, TIE, AAT, AARH, DP and SH) in 1999-2013 (15 years) are employed to train the four models (PSQRNN, BP, SVM and QRNN), and the provincial influence factors in 2014-2018 (5 years) are treated as testing data to predict the provincial EC in 2014-2018.

In our PSQRNN model, the economic factors are partially linear with EC via the parametric main effect, while economic factors and climate factors are nonlinear with EC via the ANN. Let $\vc Z=(GDP, VASI, TRSCG, TIE)$, $\vc X=(\vc Z, AAT, AARH, DP, SH)$ and $Y=EC$. The PSQRNN model is
\be\label{eq3.1}
Q_\tau(Y_{it}|\vc X_{it},\vc Z_{it},\alpha_i)=\vc Z_{it}^T\vc\beta_\tau+ANN(\vc X_{it};{\vc\theta}_\tau)+\alpha_i.
\ee
The training data is $\{(Z_{it},X_{it})\rightarrow Y_{it}, i=1,\cdots,30, t=1,\cdots,15\}$. Here and after, ``$i$" indicates the province, and ``$t$" denotes the year. ``$i=1$" is Beijing, $\cdots$, ``$i=30$" is Xinjiang, ``$t=1$" is 1999, ``$t=2$" is 2000, $\cdots$, ``$t=15$" is 2013, and so on. Then we use the testing data $\{(Z_{it},X_{it}), i=1,\cdots,30, t=16,\cdots,20\}$ to predict $\{Y_{it}, i=1, \cdots, 30, t=16, \cdots, 20\}$ by the trained PSQRNN model, to obtain $\{\hat{Y}_{it}, i=1, \cdots, 30, t=16, \cdots, 20\}$. We choose $\tau_k=0.01+0.02k$ for $k=0,\cdots,49$ so that $\tau_k\in [0.01,0.99]$, and the number of
hidden nodes in the first and second hidden layers $HL=(10,5)$. In order to evaluate
the performance of PSQRNN extensively, other number of nodes are utilized in the hidden
layers, for which, one can refer to subsection 3.3.

For BP and SVM methods, the input is $\vc X$ and output is $EC$. The procedure of BP is carried out by using the \verb"neuralnet" function in \verb"R"
package \verb"neuralnet" (Fritsch et al. \cite{Fritschetal2019}). The setting scenarios are: the backpropagation
algorithm, sum of squared errors calculation, tangent hyperbolicus-type activation function,
and the default setting. We use a two-hidden-layer network with $HL=(5,5)$, not $HL=(10,5)$ because it was overfitting. The SVM method is utilized by the \verb"svm" function in \verb"R" package \verb"e1071" (Meyer et
al. \cite{Meyeretal2019}), where the default settings are adopted, for example, the kernel used in training
and predicting is radial basis $\exp(-\gamma|u-v|^2)$ with the default $\gamma=3$, and so on. For QRNN, it is modeled as
\be
Q_\tau(Y_{it}|\vc X_{it})=ANN(\vc X_{it};{\vc\theta}_\tau)\nonumber
\ee
without semiparametric and individual effect terms in contrast to the PSQRNN model (\ref{eq3.1}), where $i=1,\cdots,30$ and $t=1,\cdots,15$. We use the \verb"qrnn2" in \verb"R" package \verb"QRNN" (Cannon \cite{Canay2011}), which is developed
to fit and predict from QRNN models with two hidden layers network. We also use the number of hidden nodes $HL=(5,5)$, not $HL=(10,5)$ because of overfitting. The quantile
probability is $\tau=0.5$, which is corresponding to the least absolute deviation regression, since it is robust to the heavy-tailed data or outliers.

From two perspectives of province and year, the mean absolute percentage error
(MAPE) and relative root mean square error (RRMSE) are utilized to be the measurement of forecasting performance (out of sample), for the 30 provinces:
\be
\mathrm{MAPE}_i=\frac{1}{5}\sum_{t=16}^{20}\left|\frac{{Y}_{it}-\hat{{Y}}_{it}}{Y_{it}}\right|, ~~
\mathrm{RRMSE}_i=\frac{\sqrt{\sum_{t=16}^{20}({Y}_{it}-\hat{{Y}}_{it})^2}}{\sqrt{\sum_{t=16}^{20}Y_{it}^2}}, i=1,\cdots,30,\nonumber
\ee
and over the 2014-2018 years:
\be
\mathrm{MAPE}_t=\frac{1}{30}\sum_{i=1}^{30}\left|\frac{{Y}_{it}-\hat{{Y}}_{it}}{Y_{it}}\right|, ~~
\mathrm{RRMSE}_t=\frac{\sqrt{\sum_{i=1}^{30}({Y}_{it}-\hat{{Y}}_{it})^2}}{\sqrt{\sum_{i=1}^{30}Y_{it}^2}}, t=16,\cdots,20,\nonumber
\ee
and
\be
\mathrm{Total\cdot MAPE}=\frac{1}{150}\sum_{i=1}^{30}\sum_{t=16}^{20}\left|\frac{{Y}_{it}-\hat{{Y}}_{it}}{Y_{it}}\right|, ~~
\mathrm{Total\cdot RRMSE}=\frac{\sqrt{\sum_{i=1}^{30}\sum_{t=16}^{20}({Y}_{it}-\hat{{Y}}_{it})^2}}{\sqrt{\sum_{i=1}^{30}\sum_{t=16}^{20}Y_{it}^2}}.\nonumber
\ee
In addition, we also calculate the means and standard deviations (Std Dev) of MAPE$_i$ and RRMSE$_i$ for $i=1,\cdots,30$.
These results are listed in Tables 2 and 3, respectively.

{\setlength{\tabcolsep}{1.6em}{
\begin{table}[!h]
\begin{center}
\caption{ Measurements of forecasting performance for PSQRNN, BP, SVM and QRNN: MAPE$_i$, RRMSE$_i$ and their means and standard deviations in Subsection 3.2}\label{table3-2-1}
\vskip 0.5cm
\tabcolsep 0.1cm
\renewcommand{\arraystretch}{1.1}
\begin{tabular}{cccccccccccc}
\hline
  &\multicolumn{2}{c} {PSQRNN}&&\multicolumn{2}{c} {BP}&&\multicolumn{2}{c} {SVM}&&\multicolumn{2}{c} {QRNN}\\
\cline{2-3} \cline{5-6}\cline{8-9}\cline{11-12}
Provinces  &MAPE&RMSE                &&  MAPE&RMSE   &&MAPE&RMSE    &&MAPE&RMSE\\
 \hline
Beijing &0.196&0.208&& \textbf{0.129}&\textbf{0.130}&& 0.324&0.344&&0.276&0.280 \\
Tianjin    &\textbf{0.147}&\textbf{0.173}&&0.313&0.372&& 0.659& 0.704&&0.402&0.518 \\
Hebei  	  &\textbf{0.074}& \textbf{0.080}&&0.092& 0.114 && 0.164&0.166&& 0.162&0.186 \\
Shanxi     & \textbf{0.044}& \textbf{0.054}&&0.199&0.211&& 0.090&0.120&&0.156&0.188 \\
Inner Mongolia	&0.238& 0.313&&\textbf{0.109}&\textbf{0.119}&& 0.136&0.146&& 0.183& 0.253\\
Liaoning &0.168& 0.229&&0.376&0.386&& 0.283&0.334&&\textbf{0.057}&\textbf{0.060}\\
Jilin &\textbf{0.116}&\textbf{0.148}&&0.288&0.327&& 0.672&0.702&&0.579&0.580\\
Heilongjiang  &0.389& 0.397&&\textbf{0.091}&\textbf{0.116}&& 0.360&0.396&&0.386&0.449\\
Shanghai	  &0.151&0.159&&\textbf{0.073}&\textbf{0.082}&& 0.143&0.175&& 0.121&0.138 \\
Jiangsu       &0.126&0.136&& 0.146& 0.158&& \textbf{0.053}&\textbf{0.058}&&0.061&0.065\\
Zhejiang  	  &\textbf{0.025}&\textbf{0.033}&& 0.056& 0.067&& 0.053&0.068&&0.210&0.285\\
Anhui      	  & \textbf{0.053}&\textbf{0.056}&&0.088&0.105&& 0.155&0.164&&0.142&0.163\\
Fujian        &\textbf{0.078}&\textbf{0.083}&&0.146&0.187&& 0.166&0.208&&0.224&0.250\\
Jiangxi    	  &0.087&0.115&&0.186&0.223&& 0.138&0.161&&\textbf{0.055}&\textbf{0.062}\\
Shandong	  &0.103&0.120&&0.127&0.148&& \textbf{0.068}&\textbf{0.075}&&0.207&0.263\\
Henan     	  &\textbf{0.029}&\textbf{0.038}&&0.330&0.362&& 0.192&0.213&&0.309& 0.312\\
Hubei      	  &\textbf{0.143}&\textbf{0.151}&& 0.224&0.254&& 0.353&0.418&&0.519&0.630\\
Hunan     	  &0.354&0.380&&\textbf{0.325}&\textbf{0.370}&& 0.421&0.503&&0.549&0.649\\
Guangdong	  &\textbf{0.030}& \textbf{0.034}&&0.041&0.062&& 0.050&0.061&&0.124&0.146\\
Guangxi	      & 0.221&0.262&&0.089&0.112&& 0.231&0.258&&\textbf{0.088}&\textbf{0.088}\\
Hainan& \textbf{0.180}&\textbf{0.200}&&1.958& 1.938&& 0.293&0.364&&0.269&0.303\\
Chongqing	  &0.345& 0.372&&0.233&0.247&& 0.291&0.355&&\textbf{0.082}&\textbf{0.091}\\
Sichuan	      & \textbf{0.165}&\textbf{0.191}&&0.234&0.266&& 0.206&0.238&&0.208&0.257\\
Guizhou	      &\textbf{0.082}& \textbf{0.127}&&0.315&0.382&& 0.185&0.191&&0.201& 0.218\\
Yunan 	      &\textbf{0.052}&\textbf{0.060}&&0.158&0.230&& 0.235&0.250&&0.169&0.208\\
Shaanxi	      &\textbf{0.203}&\textbf{0.219}&&0.203&0.254&& 0.210&0.257&&0.250& 0.395\\
Gansu	      &\textbf{0.119}&\textbf{0.129}&&0.193&0.229&& 0.151&0.195&&0.263&0.274\\
Qinghai	      &\textbf{0.061}&\textbf{0.066}&&0.494&0.555&& 0.114&0.117&&0.579& 0.592\\
Ningxia	      & 0.112&0.113&&0.122&0.157&& \textbf{0.102}&\textbf{0.109}&&0.450&0.446\\
Xinjiang	  &\textbf{0.210}&\textbf{0.247}&& 0.530& 0.545&& 0.537&0.542&&0.395&0.415\\
\hline
Mean   &\textbf{0.143}  &\textbf{0.163}&& 0.262&0.290 && 0.235  & 0.263 && 0.256& 0.292\\
(Std Dev)  &(\textbf{0.095})&(\textbf{0.104}) &&(0.343)&(0.338)&&(0.164)&(0.174)&&(0.158)&(0.174)\\
\hline
\end{tabular}
\end{center}
\end{table}}}

{\setlength{\tabcolsep}{1.6em}{
\begin{table}[!h]
\begin{center}
\caption{Measurements of forecasting performance for PSQRNN, BP, SVM and QRNN: MAPE$_t$, RRMSE$_t$, Total$\cdot$MAPE and Total$\cdot$RRMSE in Subsection 3.2.}\label{table3-2-2}
\vskip 0.5cm
\tabcolsep 0.2cm
\renewcommand{\arraystretch}{1.5}
\begin{tabular}{cccccccccccc}
\hline
  &\multicolumn{2}{c} {PSQRNN}&&\multicolumn{2}{c} {BP}&&\multicolumn{2}{c} {SVM}&&\multicolumn{2}{c} {QRNN}\\
\cline{2-3} \cline{5-6}\cline{8-9}\cline{11-12}
Years  &MAPE&RMSE                &&  MAPE&RMSE   &&MAPE&RMSE    &&MAPE&RMSE\\
 \hline
2014&\textbf{0.1173}&\textbf{0.1064}&&0.2676&0.1989&&0.1784&0.1736&& 0.1859&0.1815\\
2015&\textbf{0.1423}&\textbf{0.1295}&&0.2799&0.2283&&0.2005&0.1603&& 0.2489&0.2284\\
2016&\textbf{0.1337}&\textbf{0.1221}&&0.2936&0.2435&&0.2406&0.1983&& 0.2726&0.2425\\
2017&\textbf{0.1573}&\textbf{0.1550}&&0.2142&0.1714&&0.2986&0.2180&& 0.2849&0.2496\\
2018&\textbf{0.1661}&\textbf{0.1727}&&0.2552&0.1885&&0.2542&0.1883&& 0.2871&0.2978\\
\hline
Total&\textbf{0.1364}&\textbf{0.1451}&&0.2621& 0.2058&&0.2345&0.1905&& 0.2559&0.2483\\
\hline
\end{tabular}
\end{center}
\end{table}}}

Generally, the smaller
the MAPEs and RRMSEs are, the better the method performs. Bold values indicate the best performance. From Tables \ref{table3-2-1} and \ref{table3-2-2}, the proposed PSQRNN method performs best or being close to the best one in provinces and years based on MAPEs and RRMSEs. According to means and standard deviations of MAPEs and RRMSEs in Table \ref{table3-2-1}, it can see that our PSQRNN is more robust than the BP, SVM and QRNN. From Total$\cdot$MAPEs and Total$\cdot$RRMSEs in Table \ref{table3-2-2}, for example, 0.2621(BP)$>$0.2559(QRNN)$>$0.2345
(SVM)$>$0.1363(PSQRNN) based on Total$\cdot$MAPEs, it is also clear to find that state of the art approaches such as BP, SVM and QRNN are generally inferior ability than our PSQRNN in forecasting for this panel data. Therefore, one finds that the PSQRNN has an excellent prediction
accuracy and robustness, while the BP, SVM and QRNN are not applicable since
they did not take into account the inherent structure of the panel data.

\subsection {Model training and forecasting}
Our framework of ECF based on the influencing factors via PSQRNN model is constructed in Fig. 2(c). Recall that $\vc Z=(GDP, VASI, TRSCG, TIE)$, $\vc X=(\vc Z, AAT, AARH, DP, SH)$ and $Y=EC$ in our PSQRNN model (\ref{psqrnn}).
In the following subsection, we will train the PSQRNN model and forecast the provincial EC in China under three different scenarios. The predictive performance of the PSQRNN is evaluated again based on the first two scenarios. The efforts of the first two scenarios provide a sufficient guarantee for the third scenario to be accurately applied to prediction.

\subsubsection {Flow chat of case studies}
\begin{center}
\begin{figure}
 \vskip -4cm
 \centerline{ \includegraphics[height=17cm, width=14cm]{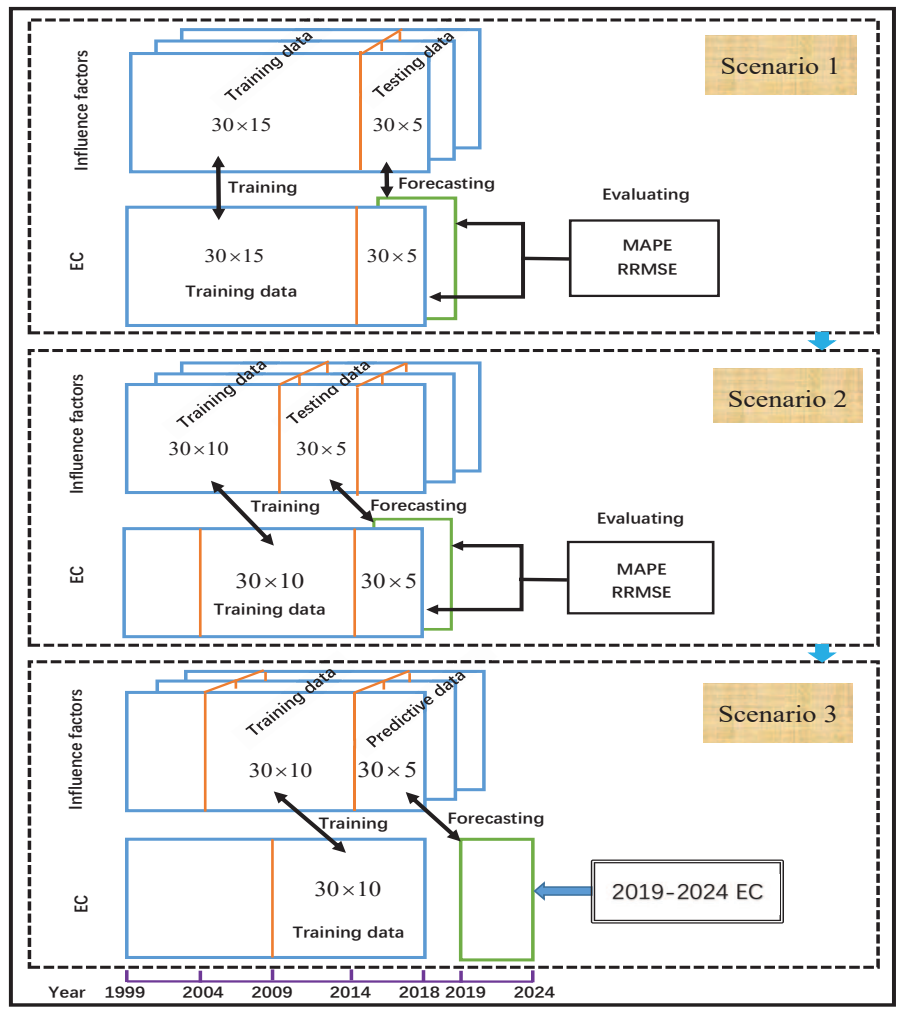}}
 \vskip -0.5in
  \caption{Flow chat of case studies.}\label{figure8}
\end{figure}
\end{center}

In order to illustrate the importance of our five-year ECF, three scenarios are considered for the five-year forecasting of EC, which is depicted in Fig. \ref{figure8}. The scenarios are:

\textbf{Scenario 1:} The provincial EC and the provincial influence factors (GDP, VASI, TRSCG, TIE, AAT, AARH, DP and SH) in 1999-2013 (15 years) are employed to train the PSQRNN model (\ref{psqrnn}), and the provincial influence factors in 2014-2018 (5 years) are treated as testing data. The scenario has been used to comparison of prediction method in Subsection 3.2. The MAPEs and RRMSEs are applied to assess the learning/predictive performance on in-sample/out-of-sample of the PSQRNN.

\textbf{Scenario 2:} The provincial EC in 2004-2013 (10 years) and the provincial influence factors in 1999-2008 (10 years) are employed to train the PSQRNN model (\ref{psqrnn}), and the provincial influence factors in 2009-2013 (5 years) are testing data to predict the provincial EC of 2014-2018 (5 years). The learning/predictive precision of the PSQRNN are also evaluated by MAPEs and RRMSEs on in-sample/out-of-sample.

\textbf{Scenario 3:} The provincial EC in 2009-2018 (10 year) and the provincial influence factors in 2004-2013 (10 years) are employed to train the PSQRNN model (\ref{psqrnn}), and the provincial influence factors in 2014-2018 are testing data to forecast EC of the next five years 2019-2023.

\subsubsection{ECF of 2014-2018 based on the setup of Scenario 1}\label{subsection3-3-2}
In Scenario 1, the PSQRNN model still is (\ref{eq3.1}). The training data is $\{(Z_{it},X_{it})\rightarrow Y_{it}, i=1,\cdots,30, t=1,\cdots,15\}$ and the testing data is $\{(Z_{it},X_{it}), i=1,\cdots,30, t=16,\cdots,20\}$. To predict $\{Y_{it}, i=1, \cdots, 30, t=16, \cdots, 20\}$ by the trained PSQRNN model, and then get $\{\hat{Y}_{it}, i=1, \cdots, 30, t=16, \cdots, 20\}$.

The setting of Scenario 1 is to evaluate the performance of the estimation and prediction. The numbers of hidden modes in the first and second layers are $HL=(5,5)$ and $HL=(10,5)$, respectively. And $\tau_k=0.01+0.02k$ for $k=0,\cdots,49$ so that $\tau_k\in [0.01,0.99]$. We compute $\mathrm{MAPE}_t$ and $\mathrm{RRMSE}_t$ for $t=1,\cdots,20$ as the evaluation criteria over 1999-2013 (in-sample) and 2014-2018 (out-of-sample). The results are listed in Table \ref{table3}. From which, the MAPEs and RRMSEs of out-of-sample are small. It shows that the PSQRNN model is well trained and the estimates are accurate. In addition, the MAPEs and RRMSEs of in-sample are $10$-$20\%$, which indicates a good forecasting performance (\cite{Dingetal2018}). Moreover, the estimation and forecasting performance tends to be better as the network nodes increasing properly.

{\setlength{\tabcolsep}{1.6em}{
\begin{table}[!h]
\begin{center}
\caption{\bf  Measurements of estimation and forecasting performance: MAPE$_t$ and RRMSE$_t$ in Subsection \ref{subsection3-3-2}}\label{table3}
\tabcolsep 0.3cm
\renewcommand{\arraystretch}{0.95}
{
\begin{tabular}{cccccc}
\hline
  &&\multicolumn{2}{c} {HL=(5,5)}&\multicolumn{2}{c} {HL=(10,5)}\\
  \cline{3-4} \cline{5-6}
&Years  &MAPE&RRMSE                &  MAPE&RRMSE\\
 \hline
in-sample
&1999&0.1886& 	0.1454& 	0.1045& 	 0.0804\\
&2000&0.1471& 	0.1032& 	0.0872& 	 0.0763\\
&2001&0.0825& 	0.0565& 	0.0409& 	 0.0335\\
&2002&0.0833& 	0.0682& 	0.0459& 	 0.0391\\
&2003&0.0809& 	0.0680& 	0.0499& 	 0.0385\\
&2004&0.0670& 	0.0616& 	0.0139& 	 0.0312\\
&2005&0.0586& 	0.0519& 	0.0342& 	 0.0456\\
&2006&0.0561& 	0.0617& 	0.0316& 	 0.0239\\
&2007&0.0760& 	0.0765& 	0.0336& 	 0.0329\\
&2008&0.0535& 	0.0443& 	0.0358& 	 0.0401\\
&2009&0.0434& 	0.0331& 	0.0312& 	 0.0326\\
&2010&0.0331& 	0.0370& 	0.0282& 	 0.0252\\
&2011&0.0553& 	0.0533& 	0.0150& 	 0.0155\\
&2012&0.0348& 	0.0319& 	0.0189& 	 0.0224\\
&2013&0.0572& 	0.0544& 	0.0227& 	 0.0258\\
\hline
out-of-sample
&2014&0.1298& 	0.1052& 	0.1173&     0.1064\\
&2015&0.1525& 	0.1321& 	0.1423& 	 0.1295\\
&2016&0.1818& 	0.1615& 	0.1337& 	 0.1221\\
&2017&0.2292& 	0.2061& 	0.1573& 	 0.1555\\
&2018&0.2290& 	0.2156& 	0.1661& 	 0.1727\\
\hline
\end{tabular}
}
\end{center}
\end{table}}}

\subsubsection{ECF of 2014-2018 based on the setup of Scenario 2}\label{subsection3-3-3}
In Scenario 2, the parametric $\vc Z=(GDP, VASI, TRSCG, TIE)$, the nonparametric $\vc X=(\vc Z, EC, AAT, AARH, DP, SH)$ and the corresponding output $Y=EC$. Based on ARIMA, PSQRNN is implemented in the following
\be\label{eq3.2}
Q_\tau(Y_{i,t+5}|\vc X_{it},\vc Z_{it},\alpha_i)=\vc Z_{it}^T\vc\beta_\tau+ANN(\vc X_{it};{\vc\theta}_\tau)+\alpha_i.
\ee
The training data is $\{(\vc Z_{it},\vc X_{it})\rightarrow Y_{i,t+5}, i=1,\cdots,30, t=1,\cdots,10\}$. Then we use the testing data $\{(\vc Z_{it}, \vc X_{it}), i=1,\cdots,30, t=11,\cdots,15\}$ to predict $\{Y_{it}, i=1, \cdots, 30, t=16, \cdots, 20\}$ by the trained PSQRNN model to obtain $\{\hat{Y}_{it}, i=1, \cdots, 30, t=16, \cdots, 20\}$. In order to forecast better, the historical EC data is used as the nonlinear predictive variable.

In PSQRNN training, we follow the same settings in Scenario 1, and calculate two evaluation criteria MAPE and RRMSE as well, which is reported in Table 5. It is obvious that the proposed model (\ref{eq3.2}) performs well for ECF.

{\setlength{\tabcolsep}{1.6em}{
\begin{table}[!h]
\begin{center}
\caption{\bf  Measurements of estimation and forecasting performance: MAPE$_t$ and RRMSE$_t$ in Subsection \ref{subsection3-3-3}}\label{table5}
\tabcolsep 0.3cm
\renewcommand{\arraystretch}{1}
{
\begin{tabular}{cccccc}
\hline
  &&\multicolumn{2}{c} {HL=(5,5)}&\multicolumn{2}{c} {HL=(10,5)}\\
  \cline{3-4} \cline{5-6}
&Years  &MAPE&RRMSE                &  MAPE&RRMSE\\
 \hline
in-sample
&2004&0.1486& 	0.1546& 	0.1581& 	 0.1540\\
&2005&0.0769& 	0.0873& 	0.0843& 	 0.0868\\
&2006&0.0361& 	0.0402& 	0.0398& 	 0.0393\\
&2007&0.0680& 	0.0793& 	0.0694& 	 0.0772\\
&2008&0.0258& 	0.0301& 	0.0241& 	 0.0286\\
&2009&0.0378& 	0.0368& 	0.0390& 	 0.0456\\
&2010&0.0267& 	0.0259& 	0.0279& 	 0.0246\\
&2011&0.0366& 	0.0366& 	0.0380& 	 0.0357\\
&2012&0.0527& 	0.0590& 	0.0504& 	 0.0588\\
&2013&0.0479& 	0.0692& 	0.0410& 	 0.0662\\
\hline
out-of-sample
&2014&0.0751& 	0.0907& 	0.0626&  0.0863\\
&2015&0.1637& 	0.1509& 	0.1526& 	 0.1412\\
&2016&0.2260& 	0.1730& 	0.2114& 	 0.1630\\
&2017&0.1920& 	0.1449& 	0.1713& 	 0.1290\\
&2018&0.1736& 	0.1360& 	0.1452& 	 0.1058\\
\hline
\end{tabular}
}
\end{center}
\end{table}}}

\subsubsection {ECF of 2019-2023 based on the setup of Scenario 3}\label{subsection3-3-4}
In Scenario 3, our goal is to predict the provincial electricity consumption of China during 2019-2023. Since there is no data of GDP, VASI, TRSCG, TIE, AAT, AARH, DP and SH for 2019-2023, model (\ref{eq3.1}) is infeasible. A historical data is utilized to predict the future electricity consumption and the candidate Model (\ref{eq3.2}) is designed as follows: the training data is $\{(\vc Z_{it},\vc X_{it})\rightarrow Y_{i,t+5}, i=1,\cdots,30, t=6,\cdots,15\}$, and the testing data is $\{(\vc Z_{it}, \vc X_{it}), i=1,\cdots,30, t=16,\cdots,20\}$ to predict $\{Y_{it}, i=1, \cdots, 30, t=21, \cdots, 25\}$ (ECF of 2019-2023). The setups are similar to Scenario 3, but there is only two hidden layers, i.e., $HL=(15,5)$ with $(\lambda_1, \lambda_2)=(0.005,0.01)$ being applied to the ANN.

\begin{center}
\begin{figure}
\centerline{ \includegraphics[height=10cm, width=18cm]{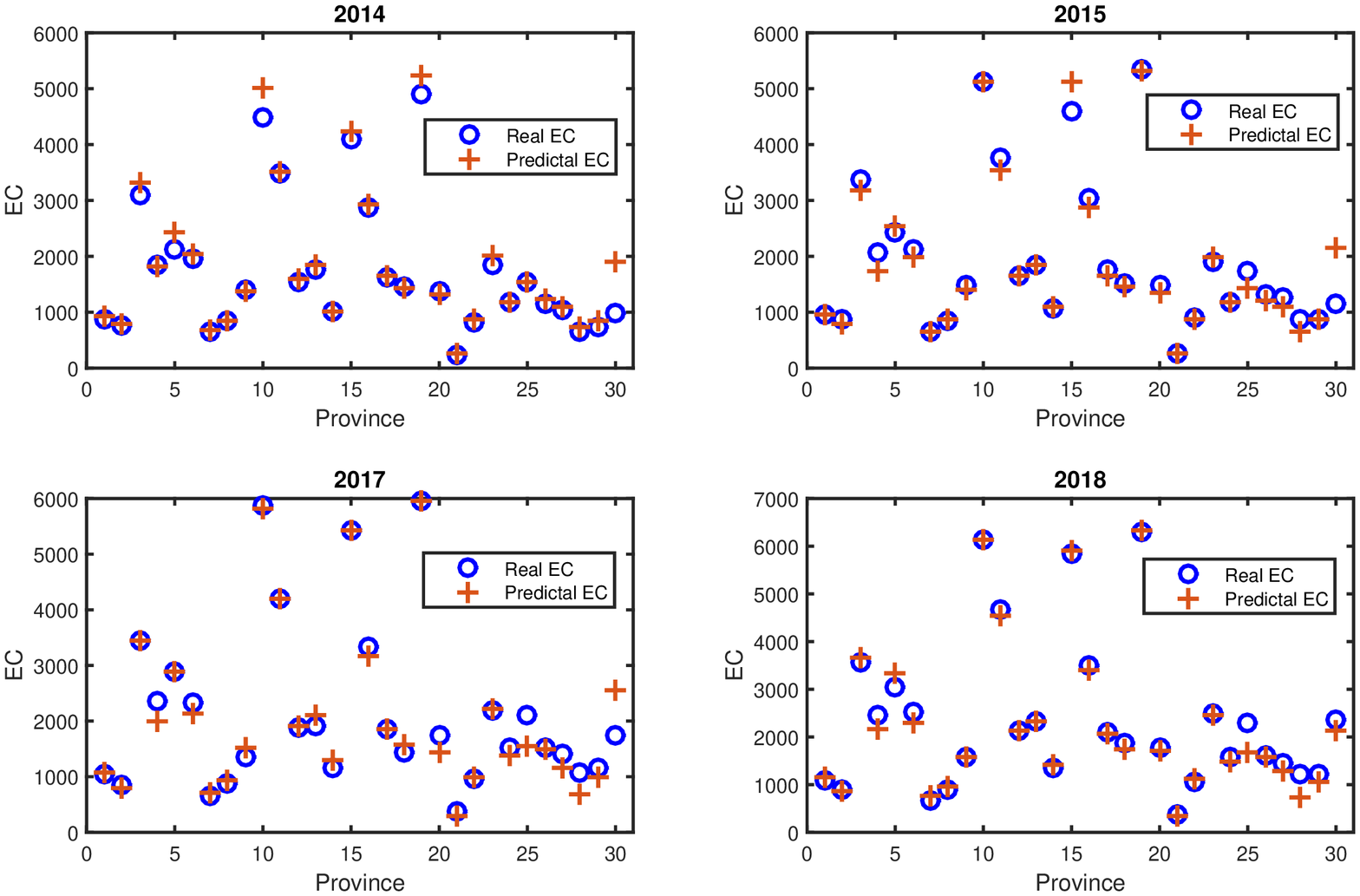}}
  \caption{Scatter plots of ECs and their predicted values in 2014-2015 and 2017-2018. Note: ``1-30" in X-axis represent the provinces, for example, ``1" is Beijing.}\label{figure9}
\end{figure}
\end{center}

\begin{center}
\begin{figure}
\centerline{\includegraphics[height=10cm, width=18cm]{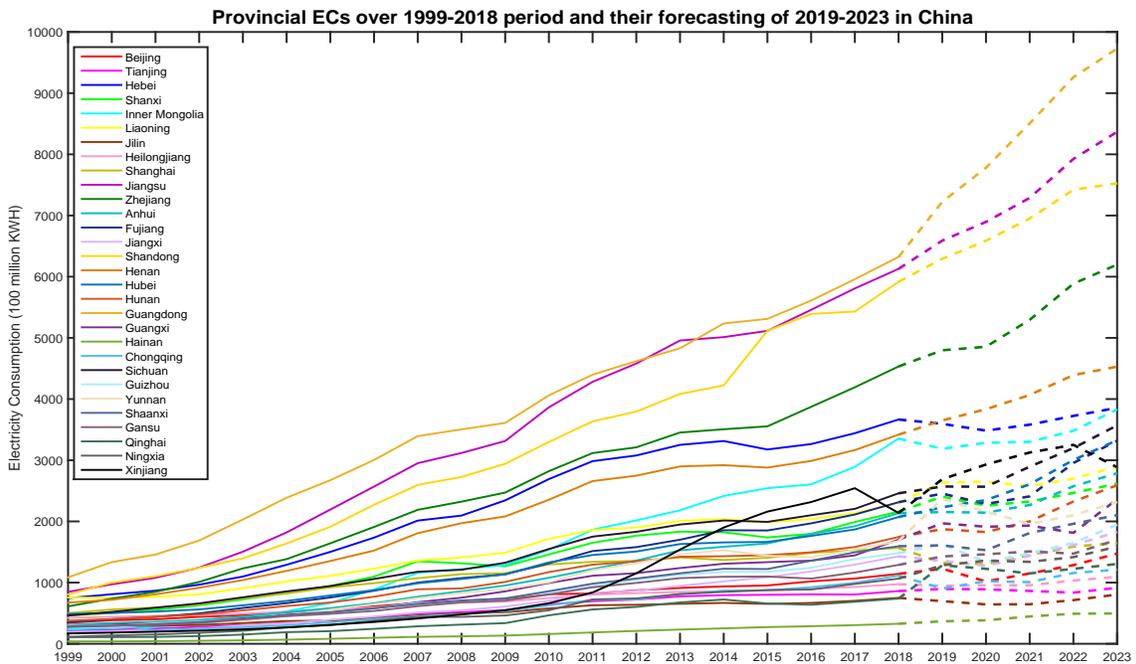}}
\vskip -0.5cm
\caption{Provincial ECs over 1999-2018 and their forecasting of 2019-2023 in China}\label{figure10}
\end{figure}
\end{center}

First of all, the performance of forecasting is evaluated. The MAPE$_t$s and RRMSE$_t$s are calculated based on $Y_{it}$ and $\hat{Y}_{it}$, $i=1,\cdots,30$ and $t=16,\cdots,20$ (Years 2014-2018), which are listed Table \ref{table6} with an excellent performance (\cite{Dingetal2018}) in terms of in-sample forecasting. Figure \ref{figure9} is a scatter plot of the ECs and the corresponding predicted provincial values of 2014-2015 and 2017-2018, which reveals an accurate prediction. Second, the estimators of the coefficients $\beta$ in model (\ref{eq3.2}) are $\hat{\beta}=(0.1119, 0.1399, 0.2928, 0.2421)$. It shows that GDP, VASI, TRSCG and TIE have a positive linear effect on EC. Finally, the ECs of 2019-2023 are predicted in Figure \ref{figure10} based on the trained PSQRNN model and the historical data. The curves in Figure \ref{figure10} are potential trends of electricity consumption demand of China's provinces in the future. The trends vary from province to province, for example, the electricity consumption demand highly in Guangdong, Jiangsu, Shandong and Zhejiang, with a relative steep growth, while the electricity demands in provinces like Sichuan, Hubei, Hunan and Anhui increase steadily and that of provinces such as Hainan, Qinghai, Ningxia, Helongjiang and Gansu grows slowly. These findings are helpful for government decision-making.

\vskip 2cm
{\setlength{\tabcolsep}{1.6em}{
\begin{table}[!h]
\begin{center}
\caption{\bf  Measurements of forecasting performance in Subsection \ref{subsection3-3-4}.}\label{table6}
\tabcolsep 0.3cm
\renewcommand{\arraystretch}{1}
{
\begin{tabular}{cccccc}
\hline
                     &2014  &2015  &2016   &2017   &2018\\
\hline
MAPE   &0.0589 &0.0778& 0.1408& 0.1091& 0.0879\\
RRMSE &0.0976 &0.1051& 0.1203& 0.0899& 0.0676\\
\hline
\end{tabular}
}
\end{center}
\end{table}}}

\newpage
\section{Conclusions}

Electricity is an important ingredient for the economic growth and development of countries. Annual electricity consumption forecasting plays a vital role in the power investment planning and energy development. However, it is particularly challenging to predict the annual electricity consumption for provinces in China, given that China has a vast territory and the electricity consumption shows great heterogeneity in provinces with different economic, social and climatic conditions. Motivated by which, we propose a new model--PSQRNN. It combines panel data, semiparametric model and composite QR with QRNN, which keeps the flexibility of nonparametric models and maintains the interpretability of parametric models simultaneously. In order to train the PSQRNN, a penalized composite quantile regression with LASSO, ridge regression and backpropagation algorithm is considered. In addition, a differentiable approximation of the quantile loss function is adopted so that the quasi-Newton optimization can be conducted to estimate the model parameters.

The prediction accuracy is evaluated by an empirical study of 30 provinces' dataset from 1999 to 2018 in China under three different scenarios, based on economic and climate factors. Compared to the QRNN model, the PSQRNN model is more robust and performs better for electricity consumption forecasting. Finally, the PSQRNN model is used to predict the electricity consumptions of provincial data over 2019-2023. We predict that China's electricity consumption will reach 9082.1 billion
KWH by 2023, which is 7.36 and 1.21 times that of 1999 and 2018, respectively. We find that the trends of electricity consumption demand in the future varies from province to province, which is helpful for government decision-making.

There are some interesting future works:
(1) For large dimensional panel data, some feature selection techniques, such as principal component analysis, factor analysis and Lasso, can be applied to our proposed PSQRNN. They can screen out more important factors from the social, economic and climatic factors associated with electricity consumption. Therefore, PSQRNN based on feature selection is worth studying for panel electricity consumption forecasting.
(2) Build hybrid methods based on Machine Learning, for example, Boosting PSQRNN, PSQRNN with random forest, PSQRNN with GANs, etc. The above methods are supposed to improve the accuracy of prediction.

\section*{Acknowledgments}

This work is supported by National Social Science Fund of China (No. 19BTJ034)


\begin{thebibliography}{99}

\bibitem{CEPPE2018} Chian Electric Power Planning and Engineering Institute. Report on China Energy Development 2018. China Electric Power Press, 2018.

\bibitem{Yuetal2015} Yu S, Wang K, Wei YM. A hybrid self-adaptive particle swarm optimization-genetic algorithm-radial basis function model for annual electricity demand prediction. Energy Conversion and Management 2015; 91:176-185.

\bibitem{Ediger2007} Ediger VS, Akar S. ARIMA forecasting of primary energy demand by fuel in Turkey. Energy Policy 2007; 35: 1701-1708.

\bibitem{Mohamedetal2010} Mohamed N, Ahmad MH, Ismail Z, Suhartono. Double seasonal ARIMA model for forecasting load demand. Matematika 2010; 26: 217-231.

\bibitem{Filik2011} Filik \"{U} B, Gerek \"{O}N, Kurban M. A novel modeling approach for hourly forecasting of long-term electric energy demand. Energy Conversion and Management 2011; 52(1):199-211.

\bibitem{Xuetal2015} Xu W, Gu R, Liu Y, Dai Y. Forecasting energy consumption using a new GMARMA model based on HP filter: the case of Guangdong Province of China. Eco Modell 2015; 45:127-135.

\bibitem{Hussainetal2016} Hussain A, Rahman M., Memon JA. Forecasting electricity consumption in Pakistan: the way forward. Energy Policy 2016; 90:73-80.

\bibitem{Cabraletal2017} Cabral JA, Legey LFL, Cabral MVF. Electricity consumption forecasting in Brazil: A spatial econometrics approach. Energy 2017; 126:124-131.

\bibitem{Oliveira2018} Oliveira EM, Oliveira FLC. Forecating mid-long term electric energy consumption through bagging ARIMA and exponential smoothing methods. Energy 2018; 144:776-788.

\bibitem{Biancoetal2013} Bianco V, Manca O, Nardini S. Linear regression models to forecast electricity consumpution in Italy. Energy Sources Part B 2013; 8:86-93.

\bibitem{MengNiu2011} Meng M, Niu D. Annual electricity consumption analysis and forecasting of China based on few observations Methods. Energy Conversion and Management 2011; 52:953-957.

\bibitem{Kaytezetal2015} Kaytez F, Taplamacioglu MC, Cam E, Hardalac F. Forecasting electricity consumption: A comparison of regression analysis, neural networks and least square support vector machines. Electrical Power and Energy Systems 2015; 67:431-438.

\bibitem{Wangetal2018} Wang J, Du P, Lu H, Yang W, Niu T. An improved grey model optimized by multi-objective and lion optimization algorithm for annual electricity consumption forecasting. Applied Soft Computing 2018; 72:321-337.

\bibitem{Dingetal2018} Ding S, Hipel KW, Dang YG. Forecasting China's electricity consumption using a new grey prediction model. Energy 2018; 149: 314-328.

\bibitem{Tangetal2019} Tang L, Wang XF, Wang XL, Shao C, Liu S, Tian S. Long-term electricity consumption forecasting based on expert prediction and fuzzy Bayesian theory. Energy 2019; 167:1144-1154.

\bibitem{Azadehetal2008a} Azaden A, Ghaderi SF, Sohrabkhani S. Annual electricity consumption forecasting by nerual network in high energy consuming industrial sectors. Energy Conversion and Management 2008; 49:2272-2278.

\bibitem{Azadehetal2008b} Azadeh A, Ghaderi SF, Sohrabkhani S. A simulated-based neural network algorithm for forecasting electrical energy consumption in Iran. Energy Policy 2008; 36:2637-2644.

\bibitem{Kandananond2011} Kandananond K. Forecasting electricity demand in Thailand with an artificial neural network approach. Energy 2011; 4:1246-1257.

\bibitem{Heetal2019} He Y, Qin Y, Wang S, Wang X, Wang C. Electricity consumption probability density forecasting method based on LASSO-Quantile Regression Neural Network. Applied Energy 2019; 233-234: 565-575.

\bibitem{PaiHong2005} Pai PF, Hong WC. Support vector machines with simulated annealing algorithms in electricity load forecasting. Energy Conversion and Management 2005; 46: 2669-2688.

\bibitem{FanChen2006} Fan S, Chen L. Short-term load forecasting based on an adaptive hybrid method. IEEE Trans Power Syst 2006; 21:392-401.

\bibitem{Hong2009} Hong WC. Electric load forecasting by support vector model. Appl Math Model 2009; 33: 2444-2454.

\bibitem{Elattaretal2010} Elattar EE, Goulermas J, Wu QH. Electric load forecasting based on locally weighted support vector regression. IEEE Trans Syst Man Cybern Part C: Appl Rev 2010; 40:438-447.

\bibitem{Yangetal2019} Yang Y, Che J, Deng C, Li L. Sequential grid approach based support vector regression for short-term electric load forecasting. Applied Energy 2019; 238:1010-1021

\bibitem{Mostafavietal2013} Mostafavi ES, Mostafavi SI, Jaafari A, Hosseinpour F. A novel machine learning approach for estimation of electricity demand: An empirical evidene from Thailand. Energy Conversion and Management 2013; 74:548-555.


\bibitem{KoenkerBassett1978} Koenker R, Bassett GJR. Regression quantiles. Econometrica1978; 46:33-50.

\bibitem{Koenker2005} Koenker R. Quantile Regression. Econometric Society Monographs 38, Cambridge Univ. Press, Cambridge, 2005.

\bibitem{Shimetal2012} Estimating value at risk with semiparametric support vector quanitle regression. Comput Stat 2012; 27:685-700.

\bibitem{Taylor2000} Taylor JW. A quantile regression neural network approach to estimating the conditional density of multiperiod returns. J. Forecast. 2000; 19 (4): 299-311.

\bibitem{Cannon2011} Cannon A. Quantile regression neural networks: implementation in R and application to precipitation downscaling. Comput. Geosci. 2011; 37 (9): 1277-1284.

\bibitem{Xuetal2016} Xu Q, Liu X, Jiang C, Yu K. Quantile autoregression neural network model with applications to evaluating value at risk. Appl. Soft Comput. 2016; 46: 1-12.

\bibitem{HeLi2018}  He Y, Li H. Probability density forecasting of wind power using quantile regression neural network and kernel density estimation. Energy Conversion and Management 2018; 164: 374-384.

\bibitem{Xuetal2017} Xu Q, Deng K, Jiang C, Sun F, Huang X. Composite quantile regression neural network with applications. Expert Syst Appl 2017; 76:129-139.

\bibitem{Cannon2018} Cannon A. Non-crossing nonlinear regression quantiles by monotone composite quantile regression neural network, with application to rainfall. Stochastic Environmental Research and Risk Assessment 2018; 32: 3207-3235.

\bibitem{Jiangetal2017} Jiang C, Jiang MJ, Xu Q, Huang X. Expectile regression neural network model with applications. Neurocomputing 2017; 247:73-86.

\bibitem{KaiLiZou2011} Kai B, Li R, Zou H. New efficient estimatiom and variable selection methods for semiparametic varying-coefficient partially linear models. The Annals of Statistics 2011; 39:305-332.


\bibitem{Engleetal1986} Engle RF, Granger CWJ, Rice J. Semiparametric estimates of the relation between weather and electricity sales. J Am Stat Assoc 1986;81:310-320.


\bibitem{FanHyndman2012} Fan S, Hyndman RJ. Short-term load forecasting based on a semi-parametric additive model. IEEE Trans Power Syst 2012; 27:134-41.

\bibitem{WeronMisiorek2008} Weron R, Misiorek A. Forecasting spot electricity prices: a comparison of parametric and semiparametric time series models. Int J Forecast 2008; 24:744-63.

\bibitem{Shaoetal2014} Shao Z, Yang SL, Gao F. Density prediction and dimensionality reduction of mid-term electricity demand in China: a new semiparametric-based additive model. Energy Convers Manage 2014; 87:439-54.

\bibitem{Goudeetal2014} Goude Y, Nedellec R, Kong N. Local short and middle term electricity load forecasting with semi-parametric additive models. IEEE Trans Smart Grid 2014; 5:440-6.

\bibitem{Shaoetal2015} Shao Z, Gao F, Zhang Q, Yang SL. Multivariate statistical and similarity measure based semiparametic modeling of the probability distribution: A novel approach to the case study of mid-log term electricity consumption forecasting in China. Applied Energy 2015; 156:502-518.

\bibitem{Lebotsaetal2018} Short term electricity demand forecasting using partially linear additive quantile regression with an application to the unit commitment problem. Applied Energy 2018; 222:104-118.

\bibitem{Caietal2018} Cai Z, Chen L, Fang Y. A semiparametric quantile panel data model with an application to estimating the growth effect of FDI. Journal of Econometrics 2018; 206:531-553.

\bibitem{Koenker2004} Koenker R. Quantile regression for longitudinal data. Journal of Multivariate Analysis 2004; 91:74-89.

\bibitem{Lamarche2010} Lamarche C. Robust penalized quantile regression estimation for panel data. Journal of Econometrics 2010; 157:396-408.

\bibitem{Galvao2011} Galvao AF. Quantile regression for dynamic panel data with fixed effects. Journal of Econometrics 2011; 164:142-157.

\bibitem{ChenLei2018} Chen W, Lei Y. The impacts of renewable energy and technological innovation on environment-energy-growth nexus: New evidence from a panel quantile regression. Renewable Energy 2018; 123:1-14.

\bibitem{WangZhuetal2018} Wang N, Zhu H, Guo Y, Peng C. The heterogeneous effect of democracy, political globalization, and urbanization on PM2.5 concentrations in G20 countries: Evidence from panel quantile regression. Journal of Cleaner Production 2018; 194:54-68.

\bibitem{WangZengLiu2019} Wang S, Zeng J, Liu X. Examining the multiple impacts of technological progress on CO$_2$ emissions in China: A panel quantile regression approach. Renewable and Sustainable Energy Reviews 2019; 103:140-150.

\bibitem{Lamarche2010} Lamarche C. Robust penalized quantile regression estimation for panel data. Journal of Econometrics 2010; 157:396-408.

\bibitem{Galvao2011} Galvao AF. Quantile regression for dynamic panel data with fixed effects. Journal of Econometrics 2011;164: 142-157.

\bibitem{Canay2011} Canay IA. A Simple Approach to Quantile Regression for Panel Data. The Econometrics Journal 2011; 14:368-386.

\bibitem{Chen2007}  Chen C. A finite smoothing algorithm for quantile regression. Journal of Computational and Graphical Statistics 2007; 16:136-164.

\bibitem{Mosteller1946} Mosteller M. On some useful ``inefficient" statistics. Ann Math Statist 1946; 17:377-408.

\bibitem {Clevertetal2016} Clevert D, Unterthiner T, Hochreiter S. Fast and accurate deep network learning by exponential linear units. ICLR, 2016.

\bibitem{Maasetal2013} Maas AL, Hannun AY, Ng AY. Rectifier nonlinearities improve neural network acoustic models. In Proceedings of the 30th International Conference on Machine Learning (ICML13), 2013.

\bibitem{Dauphinetal2014} Dauphin YN, Pascanu R, Gulcehre C, Cho K, Ganguli S, Bengio Y. Identifying and attacking the saddle point problem in high-dimensional nonconvex optimization. arXiv: 1406.2572v1, 2014.

\bibitem{Breiman1996} Breiman L. Bagging predictors. Mach Learn 1996; 24:123-140.

\bibitem{Srivastavaetal2014} Srivastava N, Hinton G, Krizhevsky A, Sutskever I, Salakhutdinov R. Dropout: A simple way to prevent neural networks from overfitting. Journal of Machine Learning Research 2014; 15:1929-1958.

\bibitem{Fanetal2019} Fan J, Hu J, Zhang X. Impacts of climate change on electricity demand in China: An empirical estimation based on panel data. Energy 2019; 170:880-888.

\bibitem{Fritschetal2019} Fritsch, S., F. Guenther, M. Wright, M. Suling, S. Mueller. 2019. R Package: neuralnet. https://github.com/bips-hb/neuralnet.

\bibitem{Meyeretal2019} Meyer, D., E. Dimitradou, K. Hornik, A. Weingessel, F. Leisch, C. Chang and C. Lin. 2019. R Package: e1071. https://cran.r-project.org/package=e1071.


\end{thebibliography}

\end{document}